\documentclass[runningheads]{llncs}

\usepackage{eccv}

\usepackage{eccvabbrv}

\usepackage{graphicx}
\graphicspath{{figures/}}
\DeclareGraphicsExtensions{.pdf,.png,.jpg}
\usepackage{booktabs}
\usepackage{algorithm2e}
\usepackage{pifont}
\newcommand{\cmark}{\ding{51}}%
\newcommand{\xmark}{\ding{55}}%
\usepackage{tikz}
\usetikzlibrary{arrows, automata, bayesnet}
\usepackage{multirow}

\newcommand{\argmax}{\mathop{\mathrm{argmax}}}
\newcommand{\EE}{\mathbb{E}}
\newcommand{\II}{\mathbb{I}}

\newcommand{\MNAME}{GPA}

\usepackage[pagebackref,breaklinks,colorlinks,citecolor=eccvblue]{hyperref}

\usepackage{orcidlink}

\begin{document}

\title{Adapting to Shifting Correlations with Unlabeled Data Calibration}

\author{Minh Nguyen\inst{1}\orcidlink{0000-0003-4762-1798} \and
Alan Q. Wang\inst{1}\orcidlink{0000-0003-0149-6055} \and
Heejong Kim\inst{2}\orcidlink{0000-0002-9871-9755} \and
Mert R. Sabuncu\inst{1,2}\orcidlink{0000-0002-7068-719X}}

\authorrunning{Nguyen et al.}

\institute{Cornell Tech, New York NY 10044, USA
\email{\{bn244,aw847,msabuncu\}@cornell.edu} \and
Weill Cornell Medicine, New York NY 10065, USA\\
\email{hek4004@med.cornell.edu}}

\maketitle

\begin{abstract}
Distribution shifts between sites can seriously degrade model performance since models are prone to exploiting unstable correlations.
Thus, many methods try to find features that are stable across sites and discard unstable features.
However, unstable features might have complementary information that, if used appropriately, could increase accuracy.
More recent methods try to adapt to unstable features at the new sites to achieve higher accuracy.
However, they make unrealistic assumptions or fail to scale to multiple confounding features.
We propose Generalized Prevalence Adjustment (\MNAME~for short), a flexible method that adjusts model predictions to the shifting correlations between prediction target and confounders to safely exploit unstable features.
\MNAME~can infer the interaction between target and confounders in new sites using unlabeled samples from those sites.
We evaluate \MNAME~on several real and synthetic datasets, and show that it outperforms competitive baselines.
\end{abstract}

\section{Introduction}\label{sec:intro}
Real-world data from multiple sites (domains) often diverge from the independent and identically distributed (i.i.d.) assumption.
Distribution shifts at different sites can cause correlations between (confounding) variables to vary significantly (\ie, instability).
For example, different hospitals may use different imaging equipment, resulting in images appearing differently. 
Some hospitals may adopt specific triaging procedures, resulting in correlations between diagnoses and other variables.
ML models often struggle to generalize to different sites (\ie, their performance does not transfer), since models are prone to shortcut learning~\cite{geirhos2020shortcut} (\ie, exploit correlations between confounding variables).
There have been several examples of shortcut learning in healthcare~\cite{albadawy2018deep,pooch2020can,degrave2021ai}.
Foundation models~\cite{radford2021learning,openai2023gpt,touvron2023llama,kirillov2023segment} trained with large amounts of diverse data can generalize out-of-the-box, in many difficult problems.
However, generalization to medical imaging is harder because the data distributions and tasks differ significantly from natural images.
Besides, training foundation models for medical imaging is also challenging, as gathering diverse data is difficult and costly.

\begin{figure}[tb]
    \centering
    \begin{subfigure}{0.48\linewidth}
        \centering
        \includegraphics[width=\linewidth]{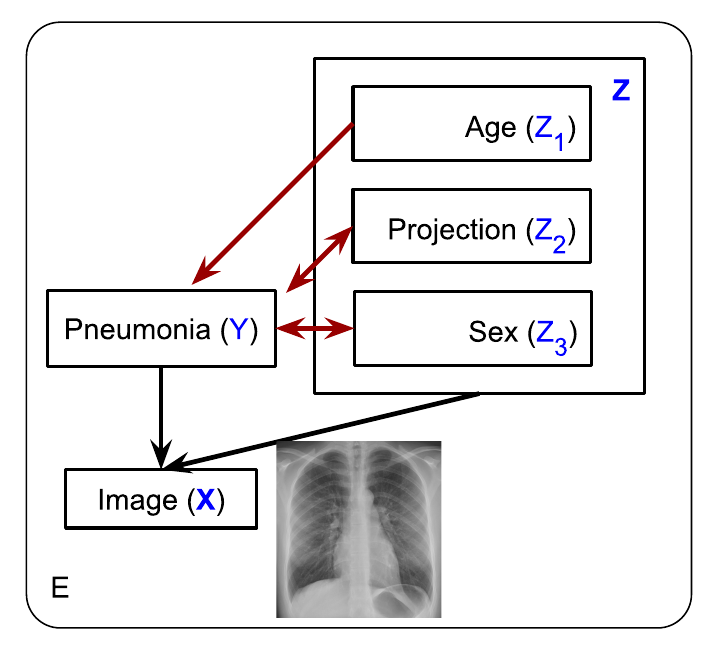}
        \caption{Anti-causal prediction task setup.}\label{fig:setup_l}
    \end{subfigure}
    \hfill
    \begin{subfigure}{0.48\linewidth}
        \centering
        \includegraphics[width=\linewidth]{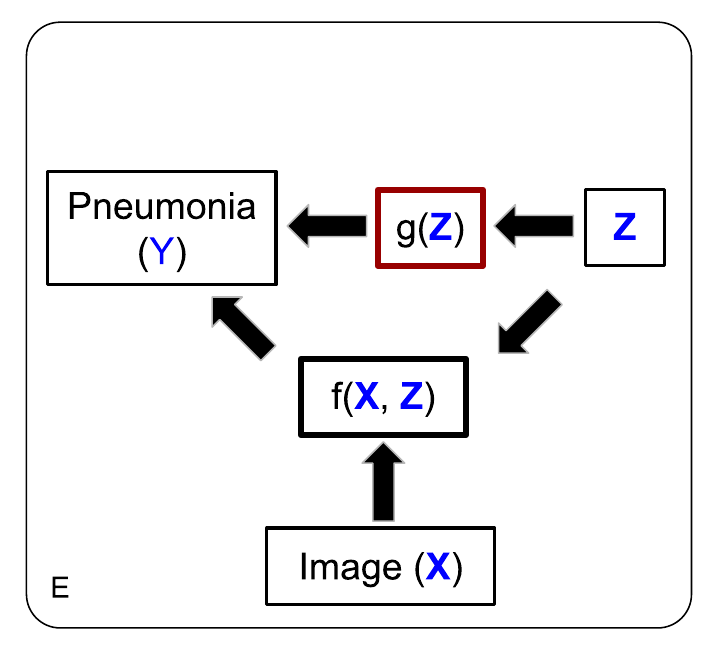}
        \caption{Modeling dataflow}\label{fig:setup_r}
    \end{subfigure}
    \caption{Left: The generative distribution of image $\mathbf{X}$ (\ie, $P(\mathbf{X}|Y,\mathbf{Z})$) is stable at different sites while the correlations between prediction target $Y$ and confounding variables $\mathbf{Z}$ can vary.
Red edges are unstable (\ie~generative mechanisms vary with sites) while black edges are stable.
Right: The generative mechanism inspires our two-part modeling approach, where one model (\ie, $f(\mathbf{X},\mathbf{Z})$) learns the stable mechanism, while another model learns the mechanism that varies with sites (\ie, $g(\mathbf{Z})$).
At a new site, only $g(\mathbf{Z})$ needs to be estimated from unlabeled data, while $f(\mathbf{X},\mathbf{Z})$ can be reused.
    }\label{fig:setup}
\end{figure}

One way to achieve out-of-domain (OOD) generalization is to avoid unstable features and only classify using stable features.
Methods such as DANN~\cite{ganin2016domain}, CORAL~\cite{sun2016deep}, and IRM~\cite{arjovsky2019invariant} ensure that no site (domain)-specific features are used for classification.
These methods assume that using solely stable features is sufficient to achieve good classification results.
However, unstable features might have complementary information, so eliminating them may lower predictive performance.
Thus, exploiting unstable features appropriately may improve both in-domain and OOD performance.

Recent adaptive methods try to exploit unstable features at new sites to achieve higher accuracy.
To exploit unstable features, it is often necessary to understand the data-generating process and where instability arises.
\Cref{fig:setup_l} shows a common anti-causal learning setup~\cite{scholkopf2012causal}, where the distribution of the input images $\mathbf{X}$ conditioned on the target label $Y$ and confounding variables $\mathbf{Z}$ is stable across sites.
However, the joint distribution of $Y$ and $\mathbf{Z}$ may vary from site to site.
TTLSA~\cite{sun2023beyond} is an adaptive method that estimates this joint distribution at new sites using unlabeled data and uses the estimated joint distribution to re-weight model predictions.
However, TTLSA requires marginalizing out the confounding variables and may not be suitable for high-dimensional or continuous $\mathbf{Z}$.
CoPA~\cite{nguyen2024robust} is another adaptive method that estimates the conditional prevalence (\ie, $P(Y|\mathbf{Z})$) instead of the joint distribution.
However, CoPA requires access to pairs of $(Y, \mathbf{Z})$ at new sites to estimate the conditional prevalence, which may be unrealistic.
Moreover, CoPA also assumes that $\mathbf{Z}$ is observed at new sites, further limiting its applicability.

We propose Generalized Prevalence Adjustment (\MNAME~for short), an adaptive method to achieve better OOD generalization in the anti-causal learning setup.
Similar to CoPA, \MNAME~also learns a stable adaptive predictor~\cite{subbaswamy2022unifying} by modeling the data-generation process using two separate estimators: one for the stable mechanism and the other for the shifting $P(Y|\mathbf{Z})$ distribution (\ie, $f$ and $g$ in \cref{fig:setup_r} respectively).
However, unlike CoPA, \MNAME~can infer the interaction between the target and the confounders at new sites using only unlabeled samples from those sites.
Specifically, \MNAME~can estimate the conditional and marginal prevalence (\ie, $P(Y|\mathbf{Z})$ and $P(Y)$) from unlabeled data at test time.
Furthermore, \MNAME~can better predict for samples without $\mathbf{Z}$ at test time even for high-dimensional $\mathbf{Z}$.
Our experiments on synthetic and real data show \MNAME~outperforming competitive baselines.

\section{Related Work}\label{sec:related}
The challenges of data distribution shifts and leveraging knowledge from one domain to improve performance in another have been addressed with several frameworks, grounded in distinct assumptions~\cite{gulrajani2021search}.
Transfer learning assumes access to labeled data from test site~\cite{wilson2020survey}.
Domain adaptation relies on the availability of unlabeled data from test sites~\cite{pan2010survey}.
Domain generalization, on the other hand, does not rely on any data from the test site~\cite{peters2016causal,arjovsky2019invariant}.
Our approach can benefit from test-site data, even though it does not require it.
Hence, our approach is more closely aligned with domain generalization.

Domain-invariant representation learning~\cite{muandet2013domain,li2018domain,zhao2019learning,tanwani2021dirl,wang2023nwirm} to find stable features across sites can benefit OOD generalization.
However, some domain-invariant learning methods may fail in the presence of label-shift~\cite{arjovsky2019invariant,zhao2019learning,tachet2020domain}.
Some notable methods for domain generalization include adversarial learning~\cite{ganin2016domain,li2018deep}, CORAL~\cite{sun2016deep}, IRM~\cite{arjovsky2019invariant}, and DRO~\cite{sagawa2019distributionally}.
IRM~\cite{arjovsky2019invariant}, which finds invariant causal predictors~\cite{peters2016causal,heinze2018invariant,nguyen2024efficient}, has been influential in tackling distribution shifts via causal understanding of data generation.
More recent methods such as IWDANN~\cite{tachet2020domain} and LAMDA~\cite{le2021lamda} try to tackle both domain adaptation and the label-shift problem.
However, these were developed for only 2 sites (1 source and 1 target) and might perform poorly in a multi-site setup~\cite{nguyen2024robust}.
Furthermore, some methods are also not applicable when the links between $Y$ are $\mathbf{Z}$ are unstable but causal~\cite{nguyen2024robust}.

Methods relying exclusively on stable features can exhibit limited performance because they ignore unstable features. 
Recent approaches have emerged to incorporate both types of features to achieve higher accuracy. 
However, these methods make certain assumptions about the data generation.
For example, SFB~\cite{eastwood2023spuriosity} (i) learns both stable and conditionally-independent unstable features; and (ii) uses the stable-feature predictions to adapt the unstable-feature predictions in the test site using unlabeled data.
SFB assumes that stable and unstable features are independent conditioned on the target.
Both CoPA~\cite{nguyen2024robust} and TTLSA~\cite{sun2023beyond} assume that the distribution of images conditioned on labels and confounders is stable.
CoPA~\cite{nguyen2024robust} models the instability in the conditional distribution $P(Y|\mathbf{Z})$ and additionally assumes that this can be estimated at test site.
TTLSA~\cite{nguyen2024robust} models the instability in the joint distribution $P(Y,\mathbf{Z})$ and estimates this distribution using unlabeled data at test time using Expectation Maximization (EM)~\cite{moon1996em}.

Saerens \etal's landmark work popularized the use EM to adapt to label-shift~\cite{saerens2002adjusting}; yet, EM may not work well if models are poorly calibrated~\cite{guo2017calibration}.
BBSL~\cite{lipton2018detecting} and RLLS~\cite{azizzadenesheli2018regularized} were proposed as alternatives that can work even when the predictions are not calibrated.
However, when calibrated properly, EM-based adaption can perform effectively~\cite{alexandari2020maximum}.

\section{Conditional Prevalence Adjustment (CoPA)}\label{sec:copa}
In this section, we briefly review the CoPA algorithm~\cite{nguyen2024robust}, as our approach will build on it.
We use uppercase letters (e.g., $\mathbf{X},Y$) to denote random variables, while lowercase letters (e.g., $y,\mathbf{z}$) will denote values.
Bolded letters (e.g., $\mathbf{X}$, $\mathbf{z}$) denote high-dimensional variables or values.
For brevity, we denote distributions at a specific site $e$, \ie~$P(\cdot | \cdot,E{=}e)$, as $P_e(\cdot | \cdot)$.

Since $P(\mathbf{X}|Y,\mathbf{Z})$ is assumed to be stable, for any two sites $a$ and $b$:
\begin{align}
    P_a(\mathbf{X} | Y,\mathbf{Z}) &= P(\mathbf{X} | Y,\mathbf{Z}) = P_b(\mathbf{X} | Y,\mathbf{Z}) \label{eq:stability}
\end{align}

By applying Bayes' rule to \Cref{eq:stability}:
\begin{align}
    &\frac{P_a(Y | \mathbf{X},\mathbf{Z}) P_a(\mathbf{X} | \mathbf{Z})}{P_a(Y | \mathbf{Z})} = \frac{P_b(Y | \mathbf{X},\mathbf{Z}) P_b(\mathbf{X} | \mathbf{Z})}{P_b(Y | \mathbf{Z})} \\
    &\Rightarrow P_b(Y | \mathbf{X},\mathbf{Z}) = P_b(Y | \mathbf{Z}) \frac{P_a(Y | \mathbf{X},\mathbf{Z})}{P_a(Y | \mathbf{Z})} \frac{P_a(\mathbf{X} | \mathbf{Z})}{P_b(\mathbf{X} | \mathbf{Z})} \;. \label{eq:bayes}
\end{align}
By collapsing all terms that do not depend on $Y$, \Cref{eq:bayes} can be rewritten as:
\begin{align}
    P_b(Y|\mathbf{X},\mathbf{Z}) = \mathsf{Norm}\bigg(P_b(Y|\mathbf{Z}) \frac{P_a(Y|\mathbf{X},\mathbf{Z})}{P_a(Y|\mathbf{Z})}\bigg) , \label{eq:prevalence}
\end{align}
where $\mathsf{Norm}$ is the normalization operation so that probabilities sum up to 1.
More detailed justification for this step can be found in Appendix~\ref{app:renorm}.
\Cref{eq:prevalence} implies that the ratio $P(Y|\mathbf{X},\mathbf{Z})/P(Y|\mathbf{Z})$ is invariant across sites.
Let $f_\theta(\mathbf{X},\mathbf{Z})$ be an estimator of this ratio parameterized by $\theta$.
Let $g^b_\phi(\mathbf{Z})$ be an estimator of $P_b(Y|\mathbf{Z})$ parameterized by $\phi$.

During training, CoPA first estimates $g^e_\phi$ using $(y, \mathbf{z})$ pairs for all training sites $e$.
Then, keeping all $g^e_\phi$ frozen, CoPA performs maximum likelihood estimation on Eq.~\eqref{eq:prevalence} with respect to $\theta$.
At a new test site $b$, CoPA first estimates $g^b_\phi(\mathbf{z})$ using $(y, \mathbf{z})$ pairs.
Then, given $(\mathbf{x}$, $\mathbf{z})$ as input, a prediction for $y$ is found by:
\begin{align}
    \widehat{y}_b(\mathbf{x},\mathbf{z}) &= \argmax_{Y} \widehat{P}_b(Y|\mathbf{x},\mathbf{z}) = \argmax \mathsf{Norm}(g^b_\phi(\mathbf{z}) f_\theta(\mathbf{x},\mathbf{z})), \label{eq:prediction}
\end{align}
where the final $\argmax$ is taken over the elements of the vector.

\section{Proposed Method}\label{sec:method}
The CoPA algorithm outlined in Section~\ref{sec:copa} makes three assumptions: (1) a stable mechanism for generating $\mathbf{X}$ from label $Y$ and confounders $\mathbf{Z}$, (2) access to confounders $\mathbf{Z}$ at training sites and novel test sites, and (3) access to the conditional prevalence $P_b(Y|\mathbf{Z})$ at novel test sites.
In this work, we extend the CoPA algorithm by relaxing these assumptions.
In Section~\ref{ssec:knockout}, we will remove assumption (2) and its necessity of $\mathbf{Z}$ at test sites by proposing to let the model learn to predict with missing $\mathbf{Z}$ using input knockout~\cite{nguyen2024knockout}.
In Section~\ref{ssec:yze_est}, we will remove assumption (3) and the necessity of $Y$ at test sites via the use of the Expectation-Maximization (EM) algorithm~\cite{moon1996em}.

We refer to this algorithm as Generalized Prevalence Adjustment (\MNAME).

\subsection{Conditional Prevalence Estimation without $Y$}\label{ssec:yze_est}
CoPA finds $g^b_\phi(\mathbf{Z})$, the estimator of $P_b(Y|\mathbf{Z})$, \emph{directly} using $(y, \mathbf{z})$ pairs from $b$.
\MNAME~estimates $g^b_\phi(\mathbf{Z})$ \emph{indirectly} using $(\mathbf{x}, \mathbf{z})$ pairs and the EM algorithm~\cite{moon1996em,saerens2002adjusting}.
Given data $\{\mathbf{x}^n, y^n, \mathbf{z}^n\}_{n=1}^N$ at a new site, the complete-data log-likelihood is:
\begin{align}
    \ell_b(\mathbf{X}, Y, \mathbf{Z}; \phi)
    &= \sum_{n=1}^N \sum_{i=1}^{|\mathcal{Y}|} \II[y^n{=}i] \log P_b^{\phi}(\mathbf{x}^n,y^n{=}i,\mathbf{z}^n) \\
    &= \sum_{n=1}^N \sum_{i=1}^{|\mathcal{Y}|} \II[y^n{=}i] \log\left\{ P_b(\mathbf{z}^n) P_b^{\phi}(y^n{=}i|\mathbf{z}^n) P(\mathbf{x}^n|y^n{=}i,\mathbf{z}^n) \right\} \\
    &= \sum_{n=1}^N \sum_{i=1}^{|\mathcal{Y}|} \II[y^n{=}i] \log P_b^{\phi}(y^n{=}i|\mathbf{z}^n) + C \;,
\end{align}
where $\II$ denotes the indicator function.
In the last step, we combine all terms which do not include $\phi$ into a constant $C$ (we drop this term hereafter). 
Since the true labels $y^n$ are unknown in our setting, we replace the indicator function with its conditional expectation, based on the latest parameter estimate $\phi^t$ at the $t$'th iteration:
\begin{align}
    &\EE\big[\II[y^n{=}i] \big| \mathbf{x}^n, \mathbf{z}^n ; \phi^t \big] = P_b^{\phi^t}(y^n{=}i|\mathbf{x}^n, \mathbf{z}^n) = \big[ \mathsf{Norm}(g^b_{\phi^t}(\mathbf{z^n}) f_\theta(\mathbf{x^n},\mathbf{z^n})) \big]_i \;,
\end{align}
where $[\cdot]_i$ denotes the $i$'th element of the vector.

At the $(t+1)$'th iteration, the objective is to maximize $Q(\phi \mid \phi^t)$ with respect to $\phi$, where $Q$ is the expected value of the log-likelihood function of $\phi$:
\begin{align}
    Q(\phi \mid \phi^t) &= \mathbb{E}_{Y \mid \mathbf{X}, \mathbf{Z}, \phi^t} [\ell_b(\mathbf{X}, Y, \mathbf{Z}; \phi)]\\
    &= \sum_{n=1}^N \sum_{i=1}^{|\mathcal{Y}|} P_b^{\phi^t}(y^n{=}i|\mathbf{x}^n,\mathbf{z}^n)  \log  P_b^{\phi}(y^n{=}i|\mathbf{z}^n) \\
    &= \sum_{n=1}^N \sum_{i=1}^{|\mathcal{Y}|} \big[ \mathsf{Norm}(g^b_{\phi^t}(\mathbf{z}^n) f_\theta(\mathbf{x}^n,\mathbf{z}^n)) \log g^b_\phi(\mathbf{z}^n) \big]_i \label{eq:em_one} \;.
\end{align}
We implement a stochastic gradient-based optimizer to maximize $Q$ with respect to $\phi$ (see \Cref{alg:em_one}).
This can be shown to be a generalized version of the EM algorithm~\cite{prescher2005tutorial} for estimating $P_b(Y|\mathbf{X})$ without direct access to $Y$.
In standard EM, the m-step is a closed-form solution.
\Cref{alg:em_one} is closer to a generalized version of EM since it uses numerical optimization in the m-step so optimum may not be guaranteed.
That said, the choice of optimizer in \Cref{alg:em_one} influences the behavior of the m-step.
When using SGD, backward() takes one gradient step, so \Cref{alg:em_one} is closer to generalized EM.
When using LBFGS, backward()fully maximizes the expectation before the next e-step, so \Cref{alg:em_one} is closer to standard EM.
Note that a similar approach can be taken to estimate $P_b(Y)$ (see \Cref{alg:em_two}).
That derivation is provided in Appendix~\ref{app:v2}.

\begin{algorithm}[tb]
\caption{$P(Y|\mathbf{Z})$ estimation at a new site using $(\mathbf{x}, \mathbf{z})$ pairs and EM.
$\odot$ denotes element-wise multiplication.
The detach() call stops gradient flow.}\label{alg:em_one}
\hrule
\KwIn{$D^b=\{(\mathbf{x}^n, \mathbf{z}^n)\}_{n=1}^N, f_\theta, T$}
\KwOut{$g^b_\phi$}
Randomly initialize $\phi$ \\
\For{$t=0,\dots,T{-}1$}{
    $L$ $\leftarrow$ 0 \\
    \For{$n=1,\dots,N$}{
        $\hat{y}$ $\leftarrow$ $\mathsf{Norm}(g^b_{\phi}(\mathbf{z}^n) f_\theta(\mathbf{x}^n,\mathbf{z}^n))$ \\
        $L$ $\leftarrow$ $L$ $-$ $\sum_i \big[ \hat{y}.\text{detach()}\odot \log g^b_\phi(\mathbf{z}^n) \big]_i$  \\
    }
    $L$.backward()
}
\hrule
\end{algorithm}

\begin{algorithm}[tb]
\caption{$P(Y)$ estimation at a new site using $(\mathbf{x})$ samples and EM.
$\odot$ denotes element-wise multiplication.
The detach() call stops gradient flow.
$\mathbf{z}_0$ is a place-holder value used by input knockout.}\label{alg:em_two}
\hrule
\KwIn{$D^b=\{\mathbf{x}^n\}_{n=1}^N, f_\theta, T$}
\KwOut{$g^b_\phi$}
Randomly initialize $\phi$ \\
\For{$t=0,\dots,T{-}1$}{
    $L$ $\leftarrow$ 0 \\
    \For{$n=1,\dots,N$}{
        $\hat{y}$ $\leftarrow$ $\mathsf{Norm}(g^b_{\phi}(\mathbf{z}_0) f_\theta(\mathbf{x}^n,\mathbf{z}_0))$ \\
        $L$ $\leftarrow$ $L$ $-$ $\sum_i \big[ \hat{y}.\text{detach()}\odot \log g^b_\phi(\mathbf{z}_0) \big]_i$  \\
    }
    $L$.backward()
}
\hrule
\end{algorithm}

\subsection{Predicting $Y$ without $\mathbf{Z}$ using Input Knockout}\label{ssec:knockout}
Using \cref{eq:prediction} to predict $Y$ requires access to $\mathbf{Z}$.
However, $\mathbf{Z}$ may be fully or partially missing at some test sites.
One way to remove this requirement is marginalizing over $\mathbf{Z}$, \ie~$P_b(Y|\mathbf{X}) = \int_{\mathbf{Z}} P_b(Y|\mathbf{X},\mathbf{Z}) P_b(\mathbf{Z}|\mathbf{X})$.
CoPA~\cite{nguyen2024robust} approximates this marginalization over $\mathbf{Z}$ by using a Monte-Carlo strategy and assuming that (1) $\mathbf{Z} \mid \mathbf{X}$ is uniformly distributed and (2) $P_b(Y)$ is a good approximation for $P_b(Y|\mathbf{Z})$.
\begin{align}
    \widehat{P}_b(Y|\mathbf{X}) &\approx \mathsf{Norm} \big( \sum_{\mathbf{Z}} \widehat{P}_b(Y|\mathbf{X},\mathbf{Z}) \big)
    \approx \mathsf{Norm} \big( \sum_{\mathbf{Z}} \mathsf{Norm}(P_b(Y) f_\theta(\mathbf{X},\mathbf{Z})) \big) \label{eq:copa_approx}
\end{align}

CoPA's approximation in \Cref{eq:copa_approx} may be intractable or may yield a poor estimate for high-dimensional $\mathbf{Z}$ or continuous $\mathbf{Z}$.
Instead, \MNAME~learns to predict $Y$ when $\mathbf{Z}$ is missing.
Specifically, during training, the values of $\mathbf{Z}$ are randomly replaced with a default value $\mathbf{z}_0$.
Then, during inference, 
\begin{align}
    \widehat{P}_b(Y|\mathbf{X}) &\approx \widehat{P}_b(Y|\mathbf{X},\mathbf{z}_0) = \mathsf{Norm}(g^b_\phi(\mathbf{z}_0) f_\theta(\mathbf{X},\mathbf{z}_0)) \label{eq:margin1} \\
    \widehat{P}_b(Y) &\approx \widehat{P}_b(Y|\mathbf{z}_0) = g^b_\phi(\mathbf{z}_0) \;, \label{eq:margin2}
\end{align}
which we refer to as marginalization with input knockout.
This ``dropout-like'' scheme to the input forces the neural networks to model conditional distributions with regard to any possible set of $Z$ variables~\cite{belghazi2019learning,ke2023neural,nguyen2022glacial,brouillard2020differentiable,lippe2021efficient,nguyen2024knockout}.
The default value $\mathbf{z}_0$ should be outside the support (range) of $\mathbf{Z}$ so that the neural networks know whether to output the full conditional distribution or the marginal distribution~\cite{yoon2018gain,li2018misgan,nguyen2024knockout}.
After the model has learned to predict with missing $\mathbf{Z}$, we can also use the model to infer the marginal distribution $P_b(Y)$ as shown in \cref{alg:em_two}.

\begin{table}[tb]
    \setlength{\tabcolsep}{3pt}
    \caption{Summary of CoPA and \MNAME~and their ablated variants.}\label{tab:method_list}
    \centering
    \begin{tabular}{lccccl}
    \toprule
    Name        & Input                      & Need $P_b(Y|\mathbf{Z})$ & Need $P_b(Y)$ & Predict w/o $\mathbf{Z}$  & Estimate $g_\phi$ \\
    \midrule
    CoPA        & $(\mathbf{X}, \mathbf{Z})$ & yes                      & no            & n/a                       & Directly \\
    CoPA$^*$    & $(\mathbf{X})$             & no                       & yes           & Monte-Carlo               & Directly \\
    GPA         & $(\mathbf{X}, \mathbf{Z})$ & no                       & no            & n/a                       & Indirectly (Alg.~\ref{alg:em_one}) \\
    GPA$^*$     & $(\mathbf{X})$             & no                       & no            & Input knockout            & Indirectly (Alg.~\ref{alg:em_two}) \\
    \bottomrule
    \end{tabular}
\end{table}

\subsection{The \MNAME~Algorithm}\label{ssec:algorithm}
\Cref{tab:method_list} lists the different instantiations of the algorithms.
CoPA$^*$ and \MNAME$^*$ are the ablated versions of CoPA and \MNAME~respectively when $\mathbf{Z}$ is withheld from the input (see Section~\ref{ssec:knockout}).
Using EM to find the maximum likelihood estimates requires the model to be well-calibrated to achieve good performance~\cite{alexandari2020maximum,garg2020unified}.
Since neural networks are often not well-calibrated~\cite{guo2017calibration}, most EM approaches relying on probability outputs from neural networks need a calibration step~\cite{garg2020unified,alexandari2020maximum,sun2023beyond}.
We calibrate $f_\theta$ using validation data similar to~\cite{sun2023beyond,eastwood2023spuriosity}.
\Cref{alg:gpa} outlines the steps in \MNAME.

\begin{algorithm}[tb]
\caption{\MNAME~algorithm}\label{alg:gpa}
\hrule
\KwIn{Data from training sites $e$, $D^e=\{(\mathbf{x}^n,y^n,\mathbf{z}^n)\}^e$}
\KwIn{Test data from new site $b$, $D^b$}
\KwOut{Predicted labels $\{y^n\}^b$}
1. Fit $g^e_\phi(\mathbf{Z})$ for each training site $e$ using $\{(y^n, \mathbf{z}^n)\}^e$ and input knockout \\
2. Fit $f_\theta(\mathbf{X},\mathbf{Z})$ using \cref{eq:prevalence} and input knockout \\
3. Calibrate $f_\theta(\mathbf{X},\mathbf{Z})$ using held-out validation data \\
4. \eIf {$D^b$ has $\mathbf{z}^n$} {
    Estimate $g^b_(\mathbf{Z})$ / $P_b(Y|\mathbf{Z})$ using \cref{alg:em_one} \\
    Predict $y^n$ using \cref{eq:prevalence} \\
}{
    Estimate $g^b_(\mathbf{z}_0)$ / $P_b(Y)$ using \cref{alg:em_two} \\
    Predict $y^n$ using \cref{eq:margin1}
}
\hrule
\end{algorithm}

\begin{table}[tb]
    \caption{Overview of GPA compared to baselines.}\label{tab:baseline_comparison}
    \centering
    \begin{tabular}{l@{\quad}c@{\quad}ccccc}
    \toprule
                                          & IRM    & DRO    & SFB    & TTLSA  & CoPA   & GPA (Ours) \\
    \midrule
    Works with single training site?      & \xmark & \cmark & \cmark & \cmark & \cmark & \cmark \\
    Works with multiple training sites?   & \cmark & \cmark & \cmark & \xmark & \cmark & \cmark \\
    Can $Z$ be continuous?                & n/a    & \xmark & n/a    & \xmark & \cmark & \cmark \\
    Can $Z$ be high-dimensional?          & n/a    & \cmark & n/a    & ?      & \cmark & \cmark \\
    Don't need $Z$ at test?               & \cmark & \cmark & \cmark & \cmark & \xmark & \cmark \\
    Don't need extra $Y$ at test?         & \cmark & \cmark & \cmark & \cmark & \xmark & \cmark \\
    \bottomrule
    \end{tabular}
\end{table}

\section{Experiments}
\subsection{Baselines}\label{ssec:baselines}
We compared \MNAME~against ERM, IRM~\cite{arjovsky2019invariant}, DANN~\cite{ganin2016domain}, CORAL~\cite{sun2016deep}, DRO~\cite{sagawa2019distributionally}, and CoPA~\cite{nguyen2024robust}.
CoPA was described above in Section~\ref{sec:copa}, and assumes that the labels $y$ are available in the new test site.
In the ablation results, $\textrm{CoPA}^*$ refers to the variant of CoPA that does not assume access to $z$ at test site and marginalizes out $\mathbf{Z}$ via Monte-Carlo sampling.
We provide a description of the baselines below:
\begin{itemize}

\item Empirical Risk Minimization (ERM) is the standard scheme that trains a neural network to predict $Y$, given $\mathbf{X}$ by minimizing cross-entropy loss on all training data and ignoring site information. 

\item 
Invariant Risk Minimization (IRM) learns representations that are invariant across different sites or environments. 
IRM aims to find features that lead to similar predictions across all sites, thus improving generalization to unseen environments.

\item
Domain-Adversarial Neural Network (DANN) trains a feature extractor to produce features that are indistinguishable between source and target domains, using a domain classifier that tries to distinguish between the two. 
The goal is to learn representations that are domain-invariant, enhancing the model's ability to generalize across domains.

\item
Correlation Alignment (CORAL) minimizes domain shift by aligning the second-order statistics (covariances) of source and target domain feature distributions. CORAL seeks to adjust the source domain features to have similar distributions to the target domain, without requiring explicit domain labels.

\item
Distributionally Robust Optimization (DRO) aims to improve the worst-case performance across a set of potential distributions, enhancing model robustness to shifts and outliers.
As unstable correlations include spurious correlations, methods designed for spurious correlations, such as IRM/DRO could be competitive.
\end{itemize}

We train a couple of ERM variants: one where $\mathbf{X}$ is the only input and another, indicated as $\textrm{ERM}^Z$, trained with $(\mathbf{x}, \mathbf{z})$ input pairs.
In the MNIST experiment, we also have a third ERM model, $\textrm{ERM}^c$ that was trained with grayscale input images, where the spurious color feature was removed.
When a baseline depends on assumptions that may be invalid in some experimental setup, we exclude that baseline from the experiment (see \Cref{tab:baseline_comparison}).
For example, when there is a single training site, IRM is not applicable.
Hence, IRM is excluded from the chest X-ray experiment.
For each method, we average results from 5 different runs using different random seeds.
Since CORAL and DANN can use unlabeled data to learn invariant representation, they were provided unlabeled data from validation and test sets.

\subsection{Implementation Details}
The $f_\theta$ and $g_\phi$ are implemented as neural networks.
For the architectural details of the networks, please see \Cref{app:extra_impl_details}.
When $g_\phi$ is optimized using EM, we used LBFGS as the optimizer and set the maximum number of EM iterations to be 5 (\ie, $T=5$ in \Cref{alg:em_one} and \Cref{alg:em_two}).
For the input knockout in \Cref{ssec:knockout}, $\mathbf{z}_0$ is chosen to be outside the range of values of $\mathbf{Z}$.
Specifically, for a categorical $Z$ variable with $|\mathcal{Z}|$ possible values, $z_0$ is $|\mathcal{Z}|+1$.
For a continuous variable $Z$, the range of value of $Z$ is remapped to be between 1 and 2 and $z_0$ is chosen to be 0 (\eg, if $Z$ lies in the range $[0, 100]$, a value of 50 is mapped to 1.5).

Two common calibration approaches that show good empirical results are bias-corrected temperature scaling (BCTS) and vector scaling (VS)~\cite{alexandari2020maximum}.
We adopt VS~\cite{alexandari2020maximum} to calibrate our model by minimizing the negative log-likelihood on the held-out data from the validation site.
The validation site data is also used to select the best model snapshot.

We use F1-score instead of accuracy to evaluate performance due to the class imbalance in the datasets.
Standard errors over these runs are indicated with error bars in the figures.

\begin{figure}[tb]
\centering
\includegraphics[width=.8\linewidth]{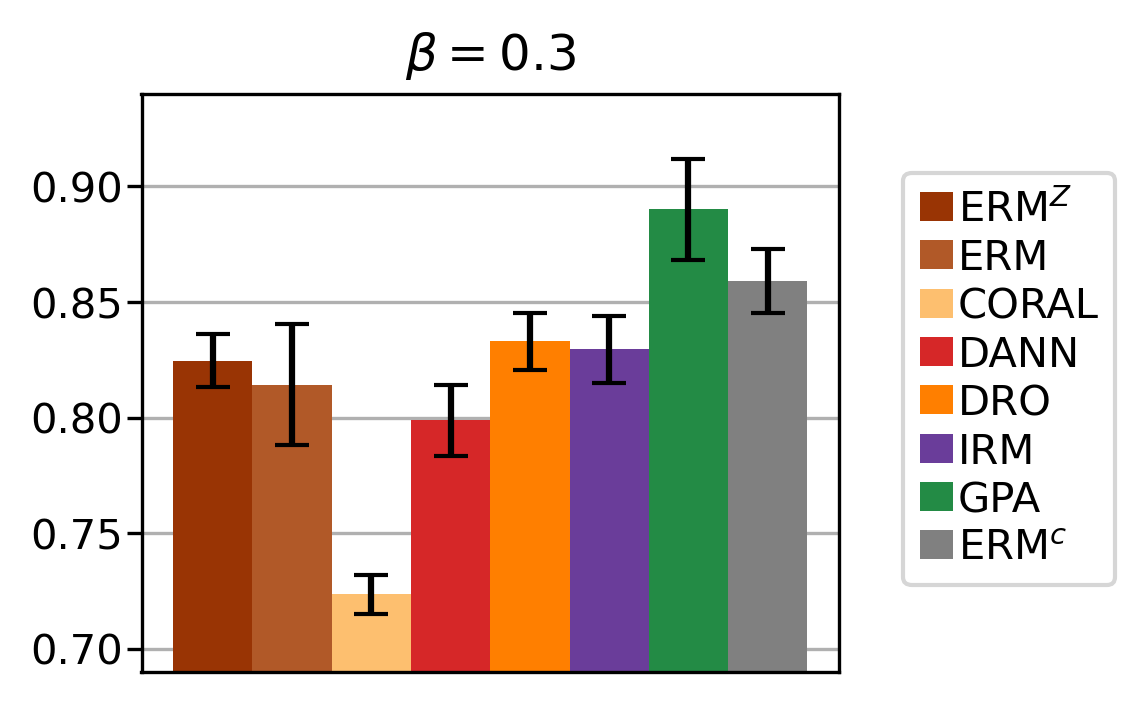}
\caption{F1-score at test site. \MNAME~outperforms all the baseline approaches.}\label{fig:cmnist_result}
\end{figure}

\subsection{Simulation - Color MNIST}\label{ssec:sim_exp}
\subsubsection{Data}
We synthesized digit images based on the MNIST dataset~\cite{lecun1998mnist}. 
The image $X$ is determined by $Y$ (randomly sampled from image set with labels $\{5, 6, 7, 8, 9\}$ when $Y=1$ and from $\{0, 1, 2, 3, 4\}$ when $Y=0$) and the color is a confounding variable $Z$ (red for $Z=1$ and green for $Z=0$). 
The $Y$ and $Z$ labels were created following Equation~\ref{eq:sem_begin}-\ref{eq:sem_end}.
Additional results with different causal relationships between $Y$ and $Z$ are shown in Appendix~\ref{app:cmnist}.
$\mathsf{Unif}(0,1)$ indicates a random variable following a uniform distribution on $(0,1)$.
The value of $\alpha$ is set at 0.3.
$\beta$ is a coefficient within the range $(0, 1)$ that represents ``site'' in real datasets. 
\begin{align}\label{eq:sem_begin}
    S &\leftarrow \mathsf{Unif}(0,1) \\
    Y &\leftarrow \mathbb{I}\big[ \beta S + (1-\beta)\alpha > 0.5 \big] \\
    Z &\leftarrow \mathbb{I}\big[ \beta S + (1-\beta)\mathsf{Unif}(0, 1) > 0.5 \big] \label{eq:sem_end}
\end{align}
Two sets, each comprising 10k samples, were used for training, generated with $\beta=0.9$ and $\beta=0.7$ respectively.
We used 0.5k independent samples created with $\beta=0.5$ for validation and 1k non-overlapping samples with $\beta=0.3$ for testing. 

\subsubsection{Results}
\begin{figure}[tb]
\centering
\includegraphics[width=.8\linewidth]{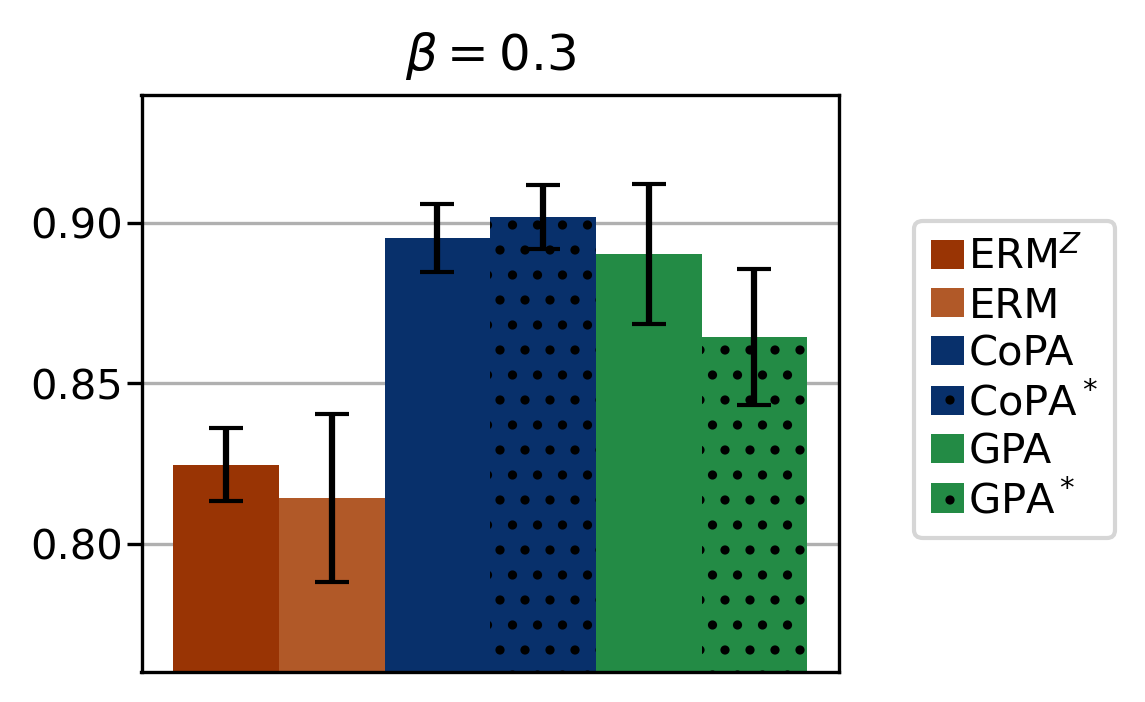}
\caption{Color MNIST experiment ablation. F1-scores are shown. $\cdot ^*$ indicates that $\mathbf{Z}$ is missing at test site for that method.}\label{fig:short}
\end{figure}

Figure~\ref{fig:cmnist_result} compares the performance of \MNAME{} against baselines, demonstrating that \MNAME{} performs the best.
Generally, we find that all domain adaptation methods (CoPA and \MNAME{}) outperform ERM.
\MNAME{} is slightly worse than CoPA, since \MNAME{} gas to indirectly estimate the prevalence, whereas CoPA has direct access to $y$'s in the test sample.
Furthermore, we compare GPA with and without $\mathbf{Z}$ at test-time by marginalizing it out with input knockout (see Fig.~\ref{fig:short}).
We observed that this decreases performance by 1-2\% as expected, but it still outperforms ERM while requiring the same amount of data.
In this simulation, $Z$ is a single binary variable and $P_b(Y)$ is a good approximation of $P_b(Y|\mathbf{Y})$, therefore CoPA$^*$'s Monte-Carlo approximation is likely reasonable.
The Monte-Carlo approximation might be less effective for multivariate $Z$ (see Section~\ref{ssec:real_exp}).

\subsection{Real Data Experiments}\label{ssec:real_exp}
\subsubsection{ISIC Data}
We replicate the experimental setup used in~\cite{nguyen2024robust}.
We used multi-center skin cancer dataset from the International Skin Imaging Collaboration (ISIC) archive, amounting 70k data samples in total~\cite{scope2009dermoscopic,gutman2016skin,codella2018skin,tschandl2018ham10000,codella2019skin,combalia2019bcn20000,rotemberg2021patient}. Each sample comprises an input image $X$, a binary target label $Y$ (melanoma or not) and confounding variables $Z$. We considered three $Z$ variables correlated with $Y$: (1) \textit{Age}, (2) \textit{Anatomical Site}, and (3) \textit{Sex}.
\textit{Anatomical Site} and \textit{Sex} are categorical variables while \textit{Age} is a continuous variable.
Since \textit{Age} is a continuous variable, methods such as DRO or TTSLA which only work for discrete grouping is not directly applicable unless the variable is discretized.
However, discretizing \textit{Age} may cause information loss that can be undesirable because \textit{Age} could be predictive of the target (as a causal parent of the target). 
We grouped the samples using spatio-temporal information (site) and split the sites into the train/validation/test sites as shown in Table~\ref{tab:isic_site}. 

\subsubsection{CXR Data}
We used the Chest X-Ray (CXR) data that consists of CXR8~\cite{wang2017chestx}, CheXpert~\cite{irvin2019chexpert}, PadChest~\cite{bustos2020padchest}, and VinDR~\cite{nguyen2022vindr}. Each sample includes an input image $X$, a binary target label $Y$ (pneumonia or not) and confounding variables $Z$. Three $Z$ variables are considered (1) \textit{Age}, (2) \textit{Projection} (AP, PA, or LL), and (3) \textit{Sex}. 
\textit{Age} is a continuous variable while \textit{Projection} and \textit{Sex} are categorical variables.
The table~\ref{tab:cxr_site} shows the train/validation/test sites along with their associated marginal prevalence.
The VinDR dataset has partially missing $\mathbf{Z}$ which can be quite common in practice since different sites may collect different amounts of meta data.
This dataset is a good benchmark for evaluating \MNAME~when $\mathbf{Z}$ is missing.
The example images of each dataset are available in the Appendix for reference.

\subsubsection{Results}
\begin{figure}[tb]
\centering
\includegraphics[width=.9\linewidth]{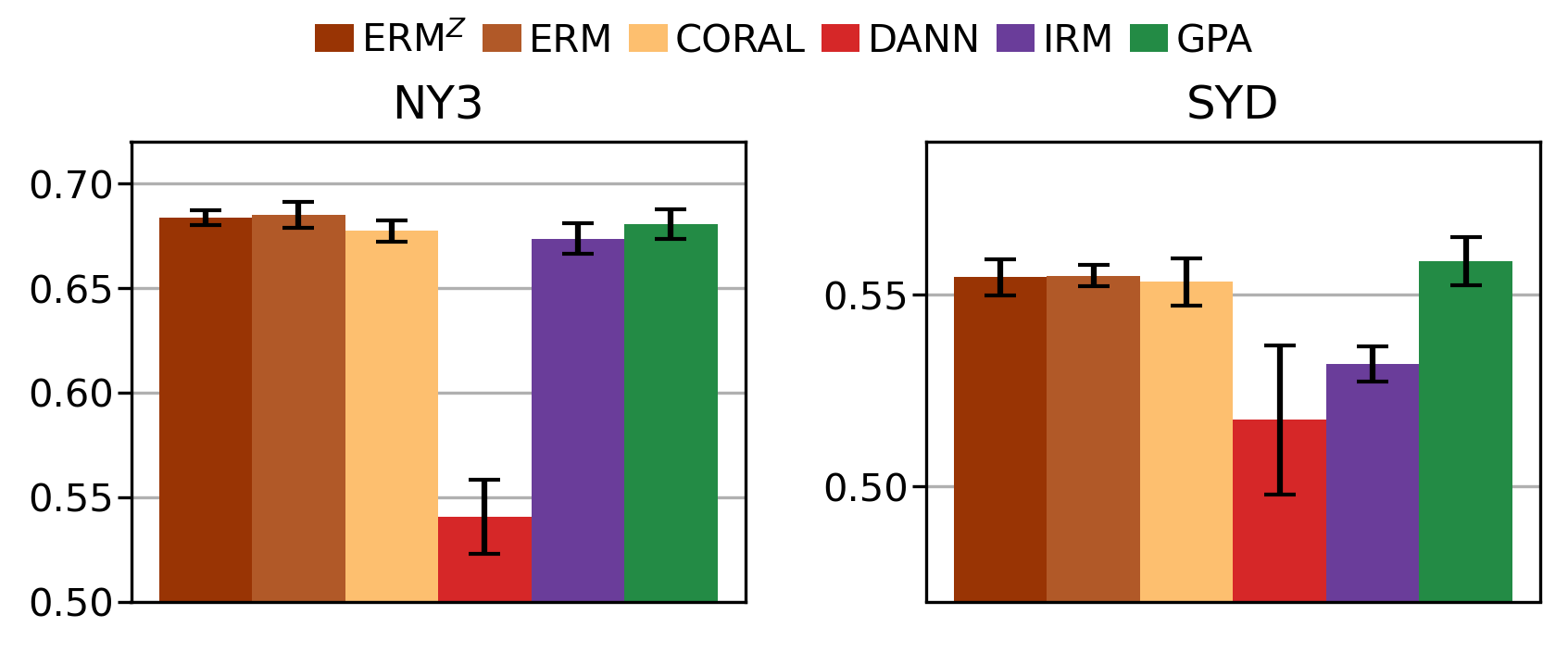}
\caption{F1-score at test sites in ISIC experiment. \MNAME~is on par or better than the baseline methods.}\label{fig:isic_main}
\end{figure}

\begin{figure}[tb]
\centering
\includegraphics[width=.9\linewidth]{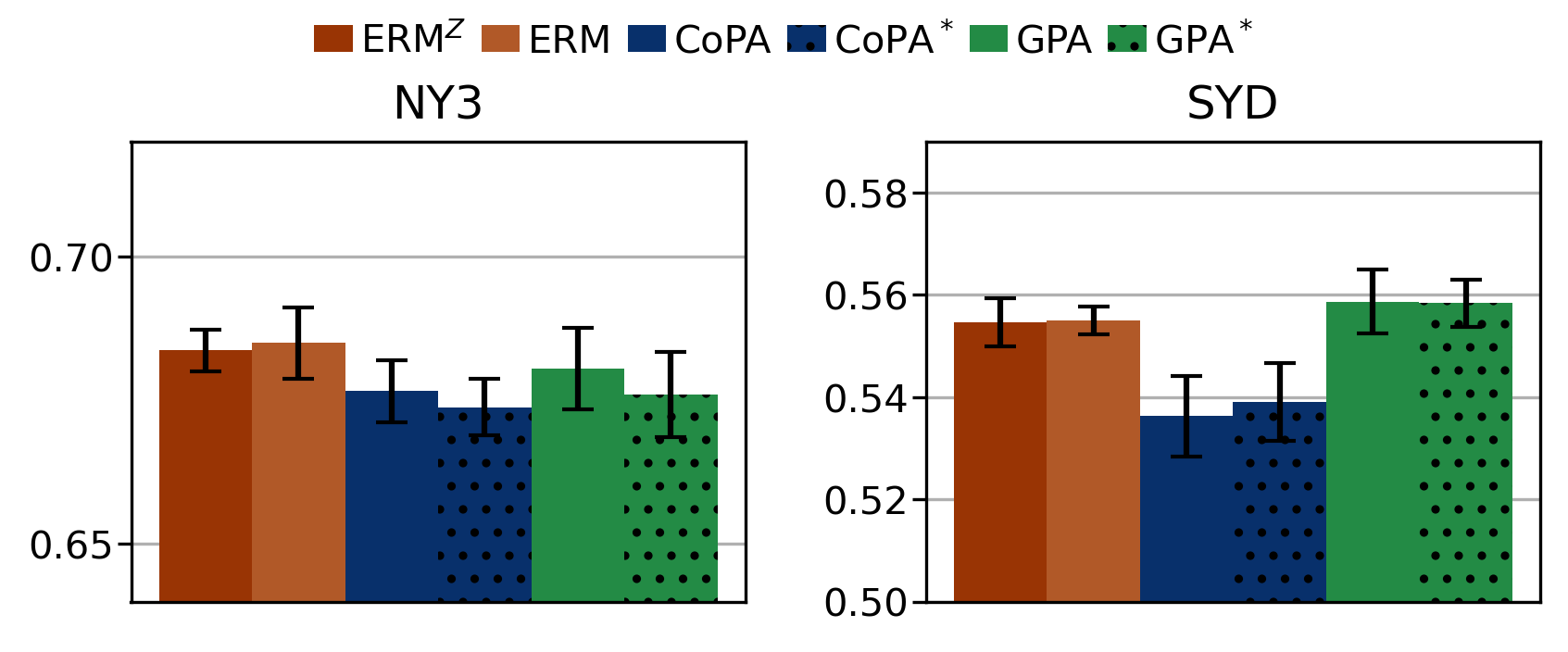}
\caption{ISIC experiment ablation. F1 scores are shown. $\cdot ^*$ indicates that $\mathbf{Z}$ is missing at test site for that method.}\label{fig:isic_supp}
\end{figure}

\begin{figure}[tb]
\centering
\includegraphics[width=.9\linewidth]{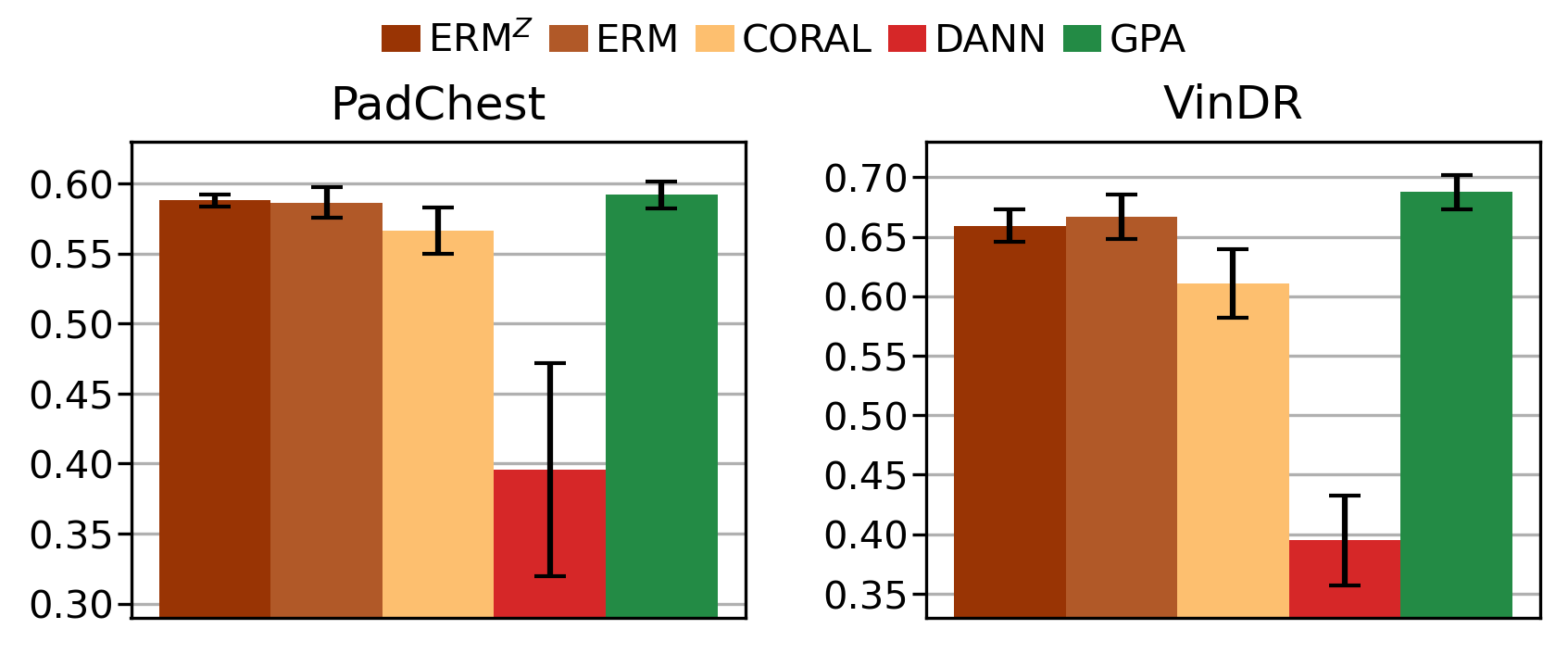}
\caption{F1-score at test sites in CXR experiment. \MNAME~outperforms the baseline methods at both test sites.}\label{fig:cxr_main}
\end{figure}

\begin{figure}[tb]
\centering
\includegraphics[width=.9\linewidth]{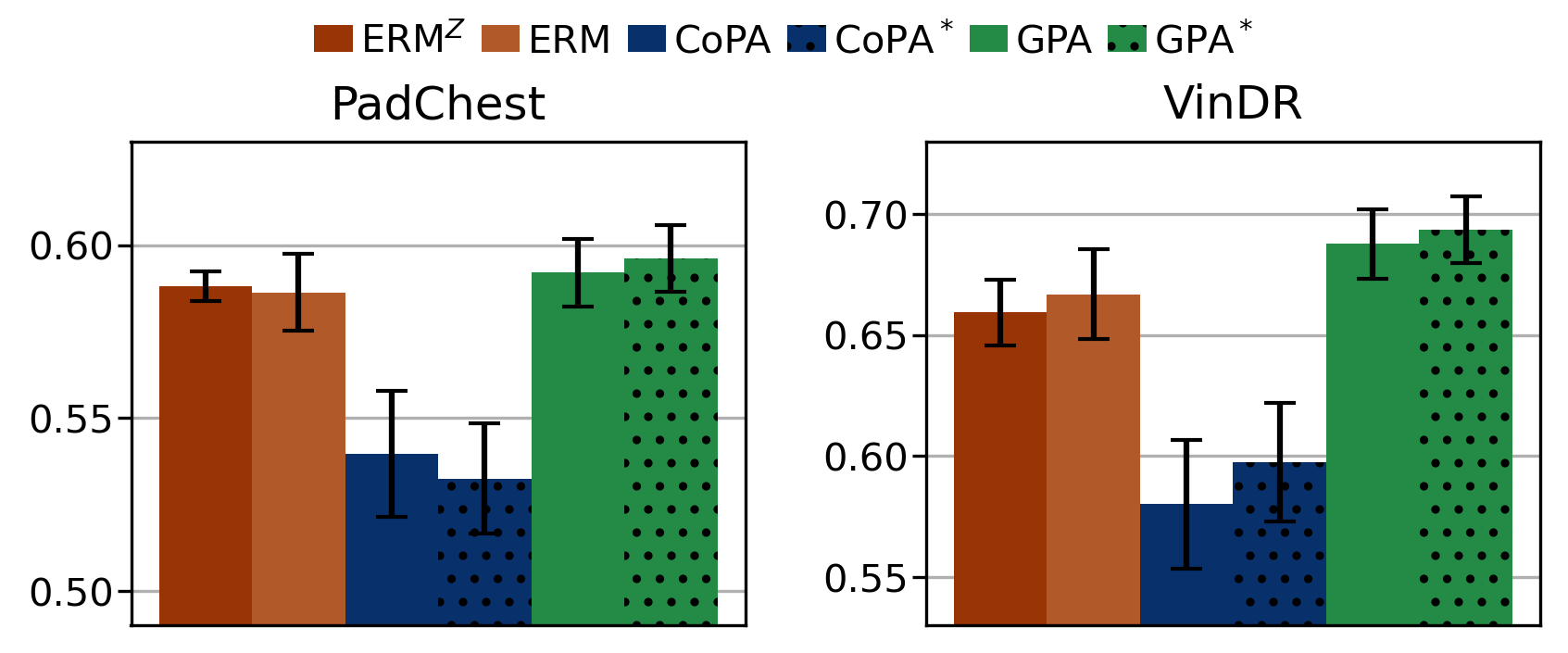}
\caption{CXR experiment ablation. F1-scores are shown of different methods when $\mathbf{Z}$ is missing.}\label{fig:cxr_supp}
\end{figure}

Fig.~\ref{fig:isic_main} shows the performance of algorithms on test sites NY3 and SYD in the ISIC dataset.
Similarly, Fig.~\ref{fig:cxr_main} shows the performance of algorithms on PadChest and VinDR datasets.
We observe that GPA using the EM algorithm and $Z$ as input is comparable to or outperforms baseline algorithms. 
Across all experiments, we find that \MNAME{} generally outperforms CoPA, indicating that the proposed EM strategy leads to better $g$ estimators and potentially less overfitting.
Within \MNAME{}, we also compare variants with and without $Z$ at adaptation time (solid vs.~textured boxes) (See Figs.~\ref{fig:isic_supp} and~\ref{fig:cxr_supp}). 
Generally, we find that using input knockout as a means to marginalize over $Z$ is an effective strategy, as the difference between variants with and without $Z$ is small. 

\begin{table}[tb]
\setlength{\tabcolsep}{4pt}
\centering
\caption{Different sites and corresponding marginal prevalence ($P(Y|E)$) in CXR.
\underline{Underlined}: validation site, \textbf{bolded}: test site.}\label{tab:cxr_site}
\begin{tabular}{lrrrrr}
    Site ($E$)              &  CXR8 & \underline{CheXpert} &   \textbf{PadChest}  & \textbf{VinDR} \\
    \midrule
    No.~of samples          & 26202 & 5886 &  4592 & 13369 \\
    \midrule
    $P(Y{=}1|E)$            & 0.049 & 0.635 & 0.082 & 0.053 \\
    \midrule
    Sex $\in\mathbf{Z}$     & Yes & Yes & Yes & Yes \\ 
    PA/AP $\in\mathbf{Z}$   & Yes & Yes & Yes & Yes \\ 
    Age $\in\mathbf{Z}$     & Yes & Yes & Yes & No  \\ 
\end{tabular}
\end{table}

\begin{table}[tb]
\setlength{\tabcolsep}{3pt}
\centering
\caption{Different sites in ISIC.
The effect of label-shift (change in $P(Y|E)$) is very pronounced between sites.
\underline{Underlined}: validation site, \textbf{bolded}: test sites.
BCN: Barcelona, MA: Massachusetts, NY: New York, QLD: Queensland, SYD: Sydney, WIE: Vienna}\label{tab:isic_site}
\begin{tabular}{lrrrrrrrrrr}
    Site ($E$)    &  BCN1 &  BCN2 &    MA &   NY1 &   \underline{NY2} &   \textbf{NY3} &   QLD &   \textbf{SYD} &  WIE1 &  WIE2 \\
    \midrule
    Samples  &  7063 &  7311 &  9251 & 11108 &  1814 &  3186 &  8449 &  1884 &  7818 &  4374 \\
    \midrule
    $P(Y{=}1|E)$ & 0.404 & 0.024 & 0.000 & 0.019 & 0.146 & 0.208 & 0.001 & 0.071 & 0.142 & 0.009
\end{tabular}
\end{table}

\section{Discussion}

In this work, we assume that the mechanism that generates $\mathbf{X}$ from $Y$ and $\mathbf{Z}$ is stable and doesn't change across sites.
This stable generation assumption requires knowing all $\mathbf{Z}$ (\ie, it must represent the sufficient adjustment set) which can cause confounding.
In the event that not all confounders are captured, it remains to be seen how this can impact performance.
One solution may be to learn some latent $\mathbf{Z}$, which is implicitly done in algorithms like IRM and more generally in the causal discovery literature.
Another limitation of the algorithm is the requirement that the full set of $\mathbf{Z}$ must be available across all sites during training.
Handling the scenario where only a subset of $\mathbf{Z}$s are available may require modifications to training, for example by drawing from ideas in the missingness literature. 

In this work, we fit separate $g$ for each environment.
This may be suboptimal in the sense that each $g$ has limited training data.
Future work might explore fitting a single model for $g$, which would allow for better support coverage at the potential cost of environment-specific adaptation.
Another question of interest is the cost of adaptation.
Future work will have to characterize how much unlabeled data is sufficient to perform well in a test site.

\section{Conclusion}
We present Generalized Prevalence Adjustment, \MNAME, which is a flexible method for OOD generalization. 
\MNAME{} adaptively adjusts model predictions to the shifting correlations between prediction target and confounders to safely exploit unstable features.
It can infer the interaction between target and confounders in new sites using unlabeled samples from those sites.
Our experimental results demonstrate the effectiveness of our model on a synthetic dataset and two real-world datasets. 

\section*{Acknowledgement}
Funding for this project was in part provided by the NIH grants R01AG053949, R01AG064027 and R01AG070988, and the NSF CAREER 1748377 grant.

\bibliographystyle{splncs04}
\bibliography{main,ref_cause,ref_crepr,ref_med}

\clearpage
\appendix

\section{Theoretical Derivations}
\subsection{CoPA/GPA Renormalization}\label{app:renorm}
Let $h(\mathbf{X},\mathbf{Z},Y) {=} P_b(Y|\mathbf{Z}) P_a(Y|\mathbf{X},\mathbf{Z}) / P_a(Y|\mathbf{Z})$.
Applying \Cref{eq:bayes} to input $(\mathbf{x}, \mathbf{z})$ yields
\begin{align}
    P_b(Y{=}i | \mathbf{X}{=}\mathbf{x},\mathbf{Z}{=}\mathbf{z}) &= h(\mathbf{x},\mathbf{z},i)
    P_a(\mathbf{X}{=}\mathbf{x}| \mathbf{Z}{=}\mathbf{z}) / P_b(\mathbf{X}{=}\mathbf{x} | \mathbf{Z}{=}\mathbf{z}) \\
    P_b(Y{=}j | \mathbf{X}{=}\mathbf{x},\mathbf{Z}{=}\mathbf{z}) &= h(\mathbf{x},\mathbf{z},j)
    P_a(\mathbf{X}{=}\mathbf{x}| \mathbf{Z}{=}\mathbf{z}) / P_b(\mathbf{X}{=}\mathbf{x} | \mathbf{Z}{=}\mathbf{z})
\end{align}
The probability vector $P_b(Y | \mathbf{X}{=}\mathbf{x},\mathbf{Z}{=}\mathbf{z})$ is the $[h(\mathbf{x},\mathbf{z},1),\dots,h(\mathbf{x},\mathbf{z},|Y|)]$ vector by scaled by the same factor $\frac{P_a(\mathbf{X}=\mathbf{x}| \mathbf{Z}=\mathbf{z})}{P_b(\mathbf{X}=\mathbf{x} | \mathbf{Z}=\mathbf{z})}$.
This scaling factor can be dropped if we renormalize the remaining terms.

\subsection{Marginal Prevalence Estimation using $X$}\label{app:v2}

Given data $\{\mathbf{x}^n, y^n\}_{n=1}^N$ at a new site, the complete-data log-likelihood is:
\begin{align}
    \ell_b(\mathbf{X}, Y ; \phi)
    &= \sum_{n=1}^N \sum_{i=1}^{|\mathcal{Y}|} \II[y^n{=}i] \log P_b^{\phi}(\mathbf{x}^n,y^n{=}i) \\
    &= \sum_{n=1}^N \sum_{i=1}^{|\mathcal{Y}|} \II[y^n{=}i] \log \left\{ P_b(\mathbf{x}^n|y^n{=}i) P_b^{\phi}(y^n{=}i) \right\} \\
    &= \sum_{n=1}^N \sum_{i=1}^{|\mathcal{Y}|} \II[y^n{=}i] \log P_b^{\phi}(y^n{=}i) + C
\end{align}
Since the true labels $y^n$ are unknown, we replace the indicator function with its conditional expectation, based on latest parameter estimate $\phi^t$ at the $t$'th iteration.
From \Cref{eq:margin1},
\begin{align}
    &\EE\big[\II[y^n{=}i] \big| \mathbf{x}^n; \phi^t \big] = P_b^{\phi^t}(y^n{=}i|\mathbf{x}^n) \approx [\mathsf{Norm}(g^b_{\phi^t}(\mathbf{z}_0) f_\theta(\mathbf{x}^n,\mathbf{z}_0))]_i , \label{eq:expectation}
\end{align}
where $[\cdot]_i$ is the $i$'th element of the vector.
At the $(t+1)$'th iteration, the objective is to maximize $Q(\phi \mid \phi^t)$ with respect to $\phi$, where $Q$ is the expected value of the log-likelihood function of $\phi$:
\begin{align}
    &Q(\phi \mid \phi^t) = \mathbb{E}_{Y \mid \mathbf{X}, \phi^t} [\ell_b(\mathbf{X}, Y; \phi)]\\
    &= \sum_{n=1}^N \sum_{i=1}^{|\mathcal{Y}|} P_b^{\phi^t}(y^n{=}i|\mathbf{x}^n) \log P_b^{\phi}(y^n{=}i) \quad\text{(from \cref{eq:expectation})} \\
    &= \sum_{n=1}^N \sum_{i=1}^{|\mathcal{Y}|} \big[ \mathsf{Norm}(g^b_{\phi^t}(\mathbf{z}_0) f_\theta(\mathbf{x}^n,\mathbf{z}_0)) \log g^b_\phi(\mathbf{z}_0) \big]_i \quad\text{(from \cref{eq:margin2})} \label{eq:em_two}
\end{align}
The marginal prevalence $g^b_\phi(\mathbf{z}_0)$ can be estimated using EM, a stochastic gradient-based optimizer, and unlabeled data (see \Cref{alg:em_two}).

\section{Additional Simulations with Color MNIST}\label{app:cmnist}
We consider two additional simulations where the relationship between $Y$ and $Z$ are causal.

\begin{figure}[t]
\centering
    \begin{subfigure}{0.48\linewidth}
    \centering
    \includegraphics[width=\linewidth]{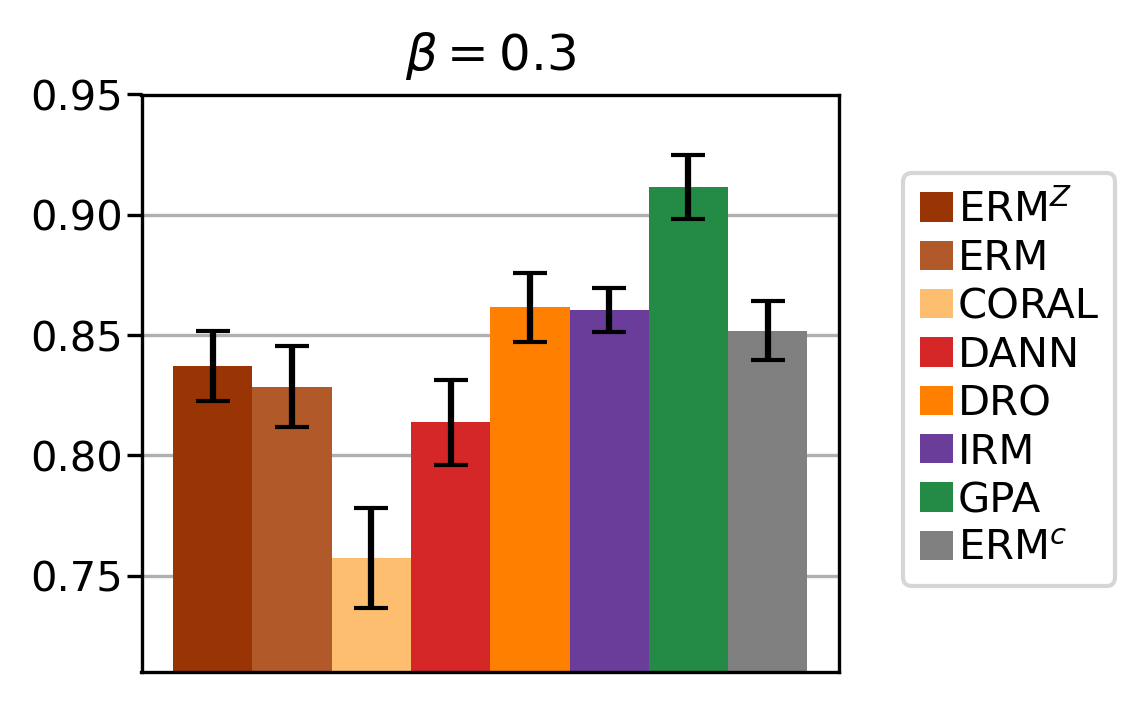}
    \caption{F1-score at test site. Y causes Z.}\label{fig:y_cause}
    \end{subfigure}
    \hfill
    \begin{subfigure}{0.48\linewidth}
    \centering
    \includegraphics[width=\linewidth]{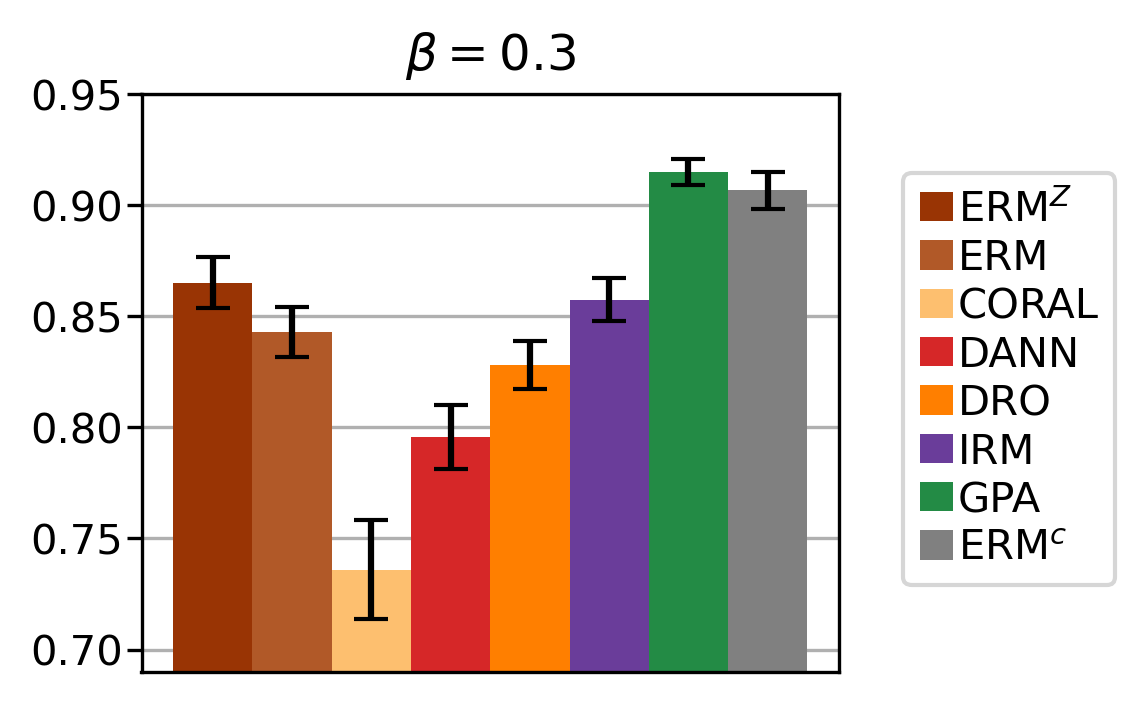}
    \caption{F1-score at test site. Z causes Y.}\label{fig:z_cause}
    \end{subfigure}
\end{figure}

$Y$ causes $Z$ (\cref{fig:y_cause})
\begin{align}
    Y &\leftarrow \mathbb{I}\big[ \beta\mathsf{Unif}(0, 1) + (1-\beta)\alpha > 0.5 \big] \\
    Z &\leftarrow \mathbb{I}\big[ \beta Y/2 + (1-\beta/2)\mathsf{Unif}(0, 1) > 0.5 \big]
\end{align}

$Z$ causes $Y$ (\cref{fig:z_cause})
\begin{align}
    Z &\leftarrow \mathbb{I}\big[ \mathsf{Unif}(0, 1) > 0.5 \big] \\
    Y &\leftarrow \mathbb{I}\big[ \beta Z/2 + \beta\mathsf{Unif}(0, 1)/2 + (1-\beta)\alpha > 0.5 \big]
\end{align}

\section{Network Architecture Details}\label{app:extra_impl_details}

\begin{figure}[h]
\centering
\includegraphics[width=\linewidth]{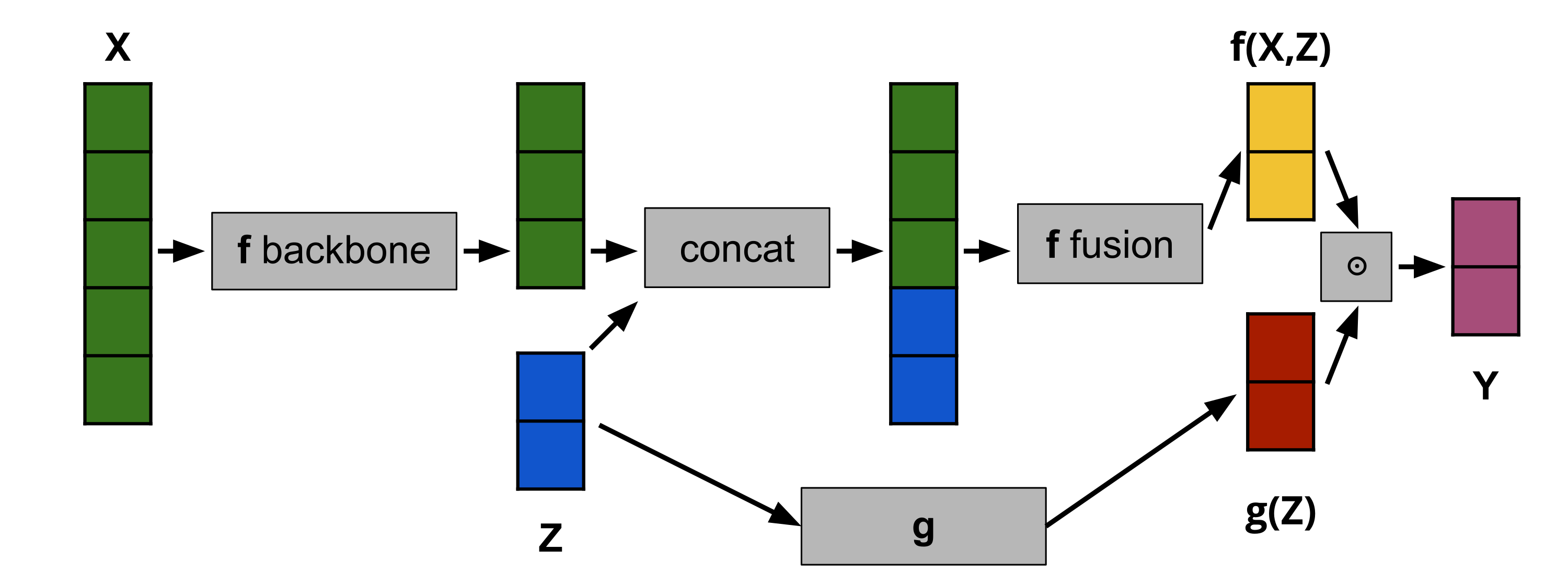}
\caption{Model architecture.}\label{fig:arch}
\end{figure}

\Cref{fig:arch} shows the general architecture of \MNAME~which consists of two networks $f_\theta$ and $g_\phi$.
The $f_\theta$ network has two modules: (1) the backbone module which reduces high-dimensional input $\mathbf{X}$ into more compact representation and (2) the fusion module which combines the representation of $\mathbf{X}$ and $\mathbf{Z}$.
\Cref{tbl:arch} outlines the specific architecture of $f_\theta$ and $g_\phi$ in the three different experimental setups.

\begin{table}[h]
\setlength{\tabcolsep}{4pt}
\centering
\caption{Neural network Architecture used in different setups.
Conv(32): convolutional layer with 32 filters,
Pool(2): max-pooling layer with size 2,
Lin(100): linear layer with output dimension of 100,
Drop(0.5): dropout layer with probability 0.5.
    }\label{tbl:arch}
\begin{tabular}{lrrr}
    Setup                   & Color MNIST & ISIC & CXR \\
    \midrule
\multirow{4}{*}{$f_\theta$ backbone} & Conv(32), ReLU, Pool(2), & ResNet50 & ResNet50 \\
                                     & Conv(32), ReLU, Pool(2), &          & \\
                                     & Conv(64), ReLU, Pool(2), &          & \\
                                     & Flatten, Lin(256)     &          & \\
    \midrule
\multirow{3}{*}{$f_\theta$ fusion}   & Lin(100), ReLU, & Lin(100), ReLU, & Lin(100), ReLU, \\
                                     & Lin(100), ReLU, & Lin(100), ReLU, & Lin(100), ReLU, \\
                                     & Lin(2)          & Lin(2)          & Lin(2)          \\
    \midrule
    $\mathbf{Z}$                     & color           & age, anat. site, sex & age, projection, sex \\
    \midrule
\multirow{5}{*}{$g_\phi$}            & Lin(100), ReLU, & Lin(100), ReLU, & Lin(100), ReLU, \\
                                     & Lin(100), ReLU, & Drop(0.5),      & Drop(0.5),      \\
                                     & Lin(2)          & Lin(100), ReLU, & Lin(100), ReLU, \\
                                     &                 & Drop(0.5),      & Drop(0.5),      \\
                                     &                 & Lin(2)          & Lin(2)          \\
\end{tabular}
\end{table}

\clearpage
\section{Chest X-Ray (CXR) Data}

\begin{figure}[ht]
\centering
\includegraphics[width=\linewidth]{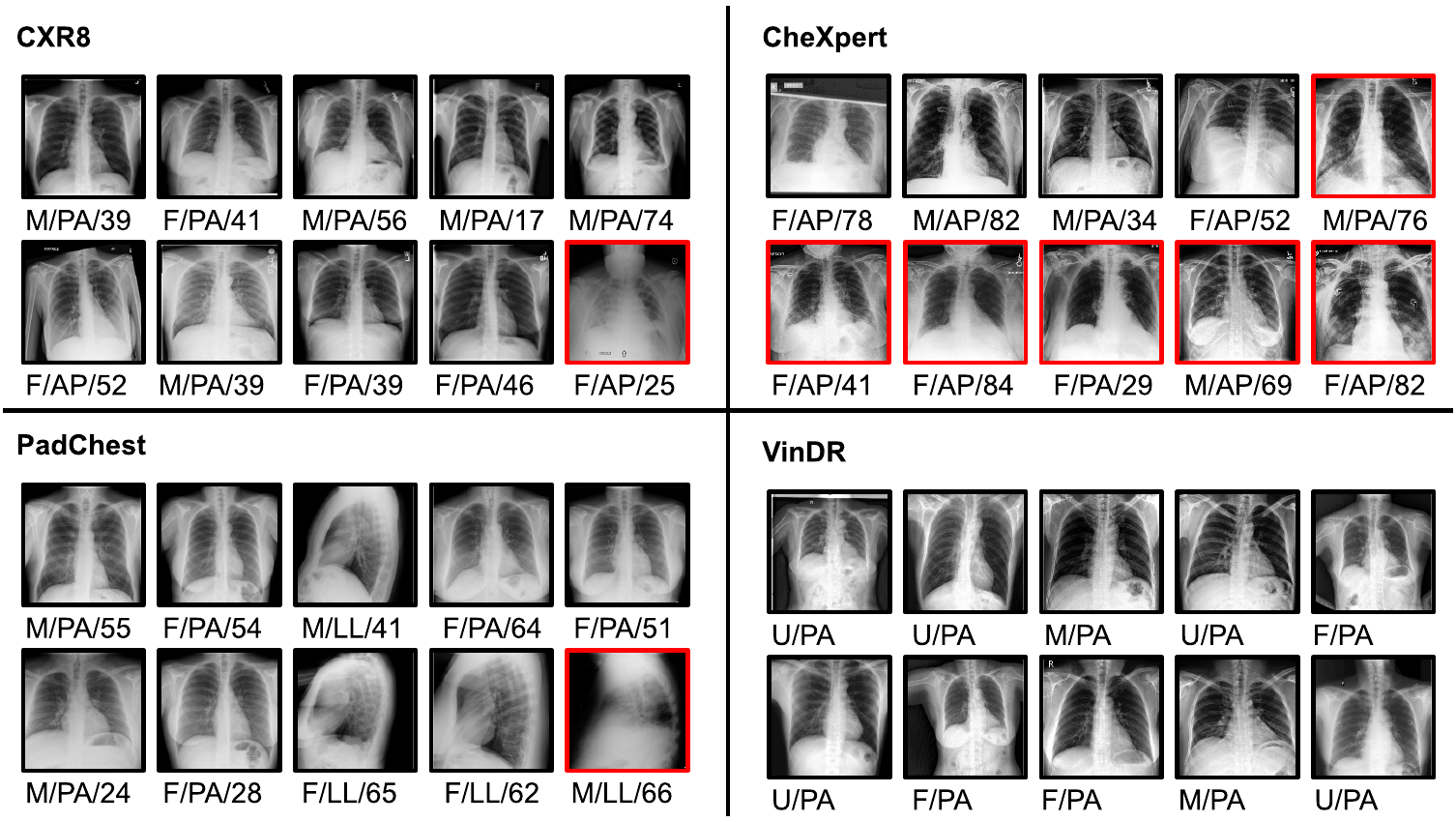}
\caption{Data from chest X-Ray (CXR) datasets used.
Red border: $Y{=}1$, black border: $Y{=}0$.
The captions indicate the \textit{Sex}/\textit{Projection}/\textit{Age} (or \textit{Sex}/\textit{Projection} for VinDR which does not have \textit{Age}).
}\label{fig:viz_cxr}
\end{figure}

\end{document}